\documentclass[runningheads]{llncs}
\usepackage{graphicx}
\usepackage{amsmath,amssymb} 
\usepackage{color}
\usepackage{subcaption}
\usepackage{tabularx}
\usepackage{multirow, makecell}
\begin{document}
\pagestyle{headings}
\mainmatter

\def\ACCV20SubNumber{872}  

\title{Depth-Adapted CNN for RGB-D cameras} 
\titlerunning{Depth-Adapted CNN for RGB-D cameras}
\authorrunning{Wu, Allibert, Stolz, Demonceaux}

\author{Zongwei Wu$^1$, Guillaume Allibert$^{2}$, Christophe Stolz$^1$, C\'edric Demonceaux$^1$ }
\institute{$^1$ VIBOT ERL CNRS 6000, ImViA, Universit\'e Bourgogne Franche-Comt\'e, France \\$^2$ Universit\'e  C\^ote d'Azur, CNRS, I3S, France 
}
\maketitle

\begin{abstract}
Conventional 2D Convolutional Neural Networks (CNN) extract features from an input image by applying linear filters. These filters compute the spatial coherence by weighting the photometric information on a fixed neighborhood without taking into account the geometric information. We tackle the problem of improving the classical RGB CNN methods by using the depth information provided by the RGB-D cameras. State-of-the-art approaches use depth as an additional channel or image (HHA) or pass from 2D CNN to 3D CNN. This paper proposes a novel and generic procedure to articulate both photometric and geometric information in  CNN architecture. The depth data is represented as a 2D offset to adapt spatial sampling locations. The new model presented is invariant to scale and rotation around the X and the Y axis of the camera coordinate system. Moreover, when depth data is constant, our model is equivalent to a regular CNN. Experiments of benchmarks validate the effectiveness of our model.
\end{abstract}

\section{Introduction}

Recent researches \cite{chen2017deeplab,Long2015FCN} prove that CNN has achieved significant progress in applications like classification, object detection, scene understanding, etc. However, the performance of 2D CNN is limited by its regular receptive field (RF) and focuses more on photometric information rather than geometry which is not directly available on RGB images. To overcome this issue, approaches such as  \cite{chen2017Atrous,Eigen2015MultiScale} modify the size of the convolution grid to contain all possible variations. Region-based CNN \cite{girshick2015fastrcnn,girshick2014rcnn,he2017maskrcnn} and their successors manage to find the Region of Interests of the object (RoI) and realize CNN on each RoI.

\begin{figure}[ht]
\begin{subfigure}{0.5\textwidth}
\includegraphics[width=\linewidth, height=4.5cm]{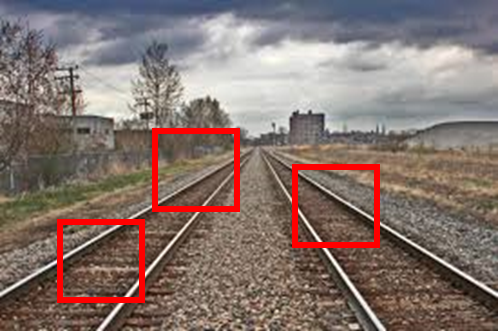} 
\caption{Conventional convolution}
\end{subfigure}
\begin{subfigure}{0.5\textwidth}
\includegraphics[width=\linewidth, height=4.5cm]{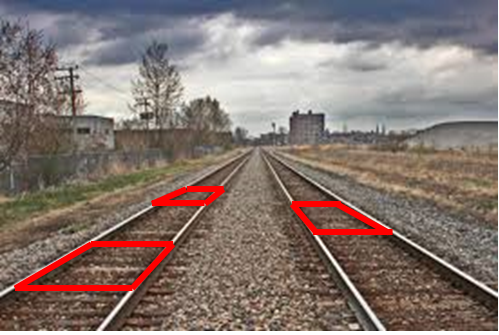}
\caption{Convolution adapted to the depth}
\end{subfigure}
\caption{Illustration of Depth-Adapted Convolution Network. The receptive field (RF) of conventional 2D CNN has a regular shape and fixed size, as shown in the left image. While with the RGB-D image, the additional cognition on depth provides the possibility to better understand RGB image. As shown in the right figure, with the depth information on the rail, we can easily link the vanishing effect with the RGB image. Inspired by this observation, we try to find a 3D planar RF whose projection on image includes more spatial information than fixed RF. In such a way, the modified 2D RF can enable progress in existing 2D CNN.}
\label{fig:image2}
\end{figure}

Recently, sensor technologies have achieved great progress in scene representation. Sensors such as Kinect, high-resolution radar, or Lidar can provide the depth map as supplementary information to RGB image. This provides the possibility to reconstruct the 3D scene with the help of both complementary modalities, which is seen as a possible improvement in CNN \cite{hazirbas2016fusenet,hu2019acnet,Lin2017RGBDCascaded,park2017rdfnet,wang2016RGBD}. In the past few years, a common approach is to take the depth map as an extra channel or extra images (HHA). While these works have proved better performance with additional depth information, variance to scale and rotation remains unsolved. In the left of Fig. \ref{fig:image2}, we can see that for two parallel rails forming the vanishing effect, the receptive fields have the same size and shape. To extract this feature in conventional CNN, either a dataset containing all variations is required, or the model should be complex enough to learn this feature.

 To overcome this issue, in this paper, we propose an end-to-end network named Depth-adapted CNN (Z-ACN). Z-ACN remains as an image convolution (2D). Instead of using a fixed receptive field, we enable an additional 2D offset to transform the shape. The new shape should be adapted to the geometry. We assume that pixels on the same 3D plane tend to share the same class. This 3D plane and depth variance have a high correlation. As illustrated in Fig. \ref{fig:image2}, we display the projection of the 3D plane of the rail on the image plane as the adapted 2D RF. It describes better the vanishing effect than a fixed RF on the left. Note that the offset is computed from the depth image with traditional computer vision algorithms that do not require gradient during back-propagation. This helps us to improve the performance of 2D CNN without complicating the model.

The main contributions of Z-ACN are summarized as follow :
\begin{itemize}
    \item We propose a generic convolutional model that adapts the 2D grid to the geometry which breaks the fixed form. This enables our model to be invariant to scale and rotation.
    \item The grid transformation is produced by traditional computer vision algorithms (without learning), which can be easily computed with minimal cost.
    \item Z-ACN can be easily integrated into any conventional CNN.
\end{itemize}


\section{Related Work}

A classic image convolution is formulated as: 
\begin{equation}
   \textbf{y}(p) = \sum_{p_n\in\textbf{R(p)}} \textbf{w}(p_n)  \cdot \textbf{x}(p + p_n).
\end{equation}
where \textbf{w} is the weight matrix. \textbf{R(p)} is the grid for point $p$. Physically it represents a local neighborhood on input feature map. \textbf{R(p)} defines the size, scale and shape of RF. For a standard $3\times3$ filter (e.g. standard regular shape with dilation $\Delta d$), the \textbf{R(p)} is given by: 
\begin{equation}
    \textbf{R(p)} = a \Vec{u} + b\Vec{v}
    \label{eq:2dgrid}
\end{equation}
where $(\Vec{u}, \Vec{v})$ is the pixel coordinate system of input feature map and $(a,b) \in (\Delta d \cdot \{-1, 0 , 1\})^2$. 
With the same input variables, different CNNs have different types of the weight matrix, grid (size, scale, shape) or pooling method. 

\subsection{Scale in 2D RGB Image Convolution}

Due to the fixed size of RF, conventional image convolution has difficulties to adapt to objects on different scales. To deal with this problem, one popular approach is to use dilated convolution \cite{chen2018parallel,yu2015Cascade}. By maintaining the kernel size but enlarging the RF, dilated convolution proves its performance on large-scale problems. Another direction is to use different scales in the network. Multi-scale approach \cite{Eigen2015MultiScale,chen2016scale1,xia2016scale2} and pyramid approach \cite{li2019pyramid,lin2017Pyramid,Zhao2017Pyramid} enable progress with CNN by considering different RF scales. But the size of these RFs is in general predefined. Moreover, as the RF shape is always regular, these methods are not eligible to deal with a non-standard object like rotated, cropped, or distorted.

\subsection{2D Deformable Model}

To learn maximum geometry information in a 2D image, researches start to insert additional transformation parameters in networks. \cite{jaderberg2015spatial} proposes a spatial transformer to align feature map. Deformable-CNN (DeformCNN) \cite{dai2017deformable} learns a dense spatial transformation to augment spatial sampling location, which breaks the regular shape of RF. \cite{Zhang2019locationdeform} adapts the DeformCNN to learn the unevenly distributed context features to improve the RoI location. Applications on videos like \cite{tian2020tdan,wang2019edvr} take consecutive frames and use DeformCNN to align the input frames to restore the video.  Nevertheless, all these methods train the offset as extra parameters during back-propagation. The objective of the deformable model is to adapt the sampling locations of CNN. This means that if the geometry properties of the camera are known in advance, it should be possible to determine the sampling position without training. One successful application is the spherical CNNs \cite{eder2019mapped,tateno2018distortion,coors2018spherenet}. These methods take advantage of the prior knowledge of image distortion to inject it explicitly into the model. In this paper, we will present how we extend this idea to another geometry property, the depth.


\subsection{CNN for 3D Representation}

Rich information on geometry provides various methods to realize 3D convolution. The volumetric representation \cite{3DConvLanding,maturana2015voxnet,wu20153d} feeds voxel data into 3D CNN. It seems to be a trivial method to deal with 3D data. However, as data is often sparse on the 3D scene, it may waste huge memory consumption for less useful information. Different from the voxel representation, \cite{Qi2017pointnet,NIPS2017Pointnet} propose to use directly the point cloud representation. Different 3D CNN methods are trying to adapt to the irregularity of point cloud. \cite{li2018pointcnn} integrates a x-transformation to leverage the spatially-local correlation of point cloud\cite{Thomas2019kpconv} introduces a spatially deformable convolution based on kernel points to study the local geometry.   \cite{liu2019relation} learns the mapping from geometry relations to high-level relations between points to get a shape awareness. \cite{liu2019densepoint} defines convolution as an SLP (Single-Layer Perceptron) with a nonlinear activator.

Some reaches also try to reduce the model complexity. \cite{tchapmi2017segcloud} adapts CRF (Conditional Random Fields) to reduce the model parameters. Multi-view method \cite{Chen_2017MultiView,Ge20173Dpose,3DFCN,Qi2016VolMultiView} reforms 3D CNN to become the combination of 2D CNNs. \cite{Chen_2017MultiView} profits from Lidar to get bird-view and front-view information in addition to a traditional RGB image. \cite{Ge20173Dpose} uses depth image to generate the 3D volumetric representation after which projections on x,y,z planes are learned respectively by 2D CNN.  3D CNN achieves better results than RGB CNN but requires further development on problems such as memory cost, data resolution, and computing time.


\subsection{RGB-D Representation}

Different from voxel and point cloud, RGB-D can profit from its image-like form and contain both photometric and geometry information. Early CNNs on RGB-D images commonly follow 2 directions. One direction is to use a depth map to create 3D point clouds where the spatial information is learned \cite{Song2016shape,song2017semantic}. This shares the same disadvantages on memory and computation cost. Another direction is to realize two separate convolutions on both RGB image and depth map and then apply a fusion in the network \cite{Lin2017RGBDCascaded,wang2016RGBD}. Some works \cite{Long2015FCN,hazirbas2016fusenet,hu2019acnet,park2017rdfnet} encode depth map to HHA image which have the same dimension of RGB image. This doubles the number of parameters and does not solve the problem of fixed size and shape of RF.

Recent works begin to adapt depth information in the convolution of the RGB image. Frustum methods \cite{qi2018frustum,Tang2019frustumplus} compute 2D RoI on RGB image and back-project them to 3D with the help of depth, which avoids the problem of regular shape. However, a huge part of computation has been done with the point cloud, which joins the disadvantages of 3D CNN. \cite{Wang2018DCNN} analyzes the depth similarity to adjust the weight for a conventional RF.  \cite{chu2018surfconv} projects 3D convolution on the 2D image which can adjust the size of RF to the depth. But for both methods, the RF shape remains regular. 

In this paper, we present a different vision to integrate depth information. Inspired by the idea of deformable convolution \cite{dai2017deformable}, our method introduces an offset into the basic operations of CNN that breaks the limitation of the fixed structure. Different from state-of-the-art methods that train the offset as a variable in the network, our offset is computed directly from depth with traditional computer vision algorithms. Thus, our model does not add extra parameters needing to be learned. It can be easily integrated into any existing model by replacing simply convolution by guided deformable convolution. The final convolutional network remains 2D but with further geometric information.

\section{Depth-Adapted Convolution Network}

A 2D convolutional grid adapted to the depth information is the prospective topic in computer vision. Different from the conventional convolution Eq. \ref{eq:2dgrid}, the Z-ACN is presented as: 
\begin{equation}
   \textbf{y}(p) = \sum_{p_n\in\textbf{R(p)}} \textbf{w}(p_n)  \cdot \textbf{x}(p + p_n + \Delta p_n). 
   \label{eq}
\end{equation}

The convolution may be operated on the irregular positions $p_n+\Delta p_n$ as the offset $\Delta p_n$ may be fractional. To address the issue, we use the bilinear interpolation which is the same as that proposed in \cite{dai2017deformable}.

The model requires 2 inputs: input feature map and depth map. The feature map is denoted as $\textbf{x} \in \mathbf{R}^{c_{i}\times h\times w}$, where $c_i$ is the number of input feature channel, $h$ and $w$ are the height and weight of the input feature map. The depth map is denoted as $\textbf{D} \in \mathbf{R}^{h\times w}$. $\textbf{D}$ is used to adapt the spatial sampling locations by computing the offset, denoted as $\Delta p \in \mathbf{R}^{c_{off}\times h_1\times w_1}$, where $h_1$ and $w_1$ are the height and weight of the output feature map and $c_{off} = 2\times N \times N$ for a $N\times N$ filter. Different from DCNN, our offset does not require gradient during back-propagation. The output feature map is denoted as $\textbf{y} \in \mathbf{R}^{c_{o}\times h_1\times w_1}$, where $c_{o}$ is the number of output feature channels.

 In the following parts, we will explain how the Z-ACN works to compute the 2D offset from depth. 

\begin{figure}[ht]
\begin{subfigure}{0.24\textwidth}
\includegraphics[width=\linewidth]{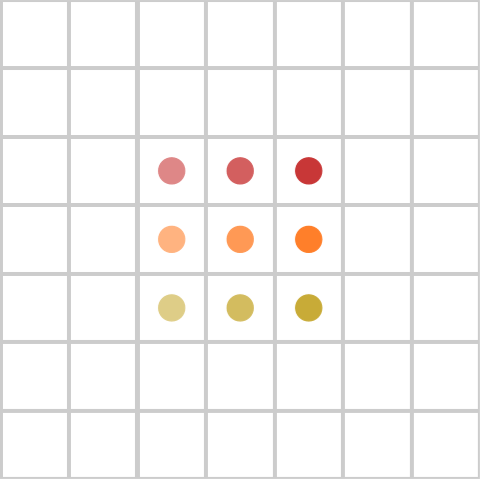} 
\caption{Standard}
\end{subfigure}
\begin{subfigure}{0.24\textwidth}
\includegraphics[width=\linewidth]{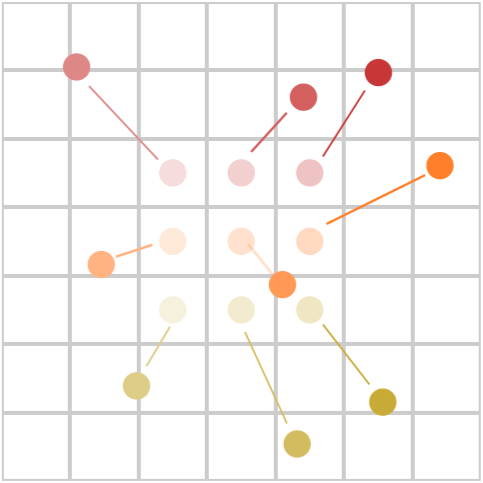}
\caption{Deformable}
\end{subfigure}
\begin{subfigure}{0.24\textwidth}
\includegraphics[width=\linewidth]{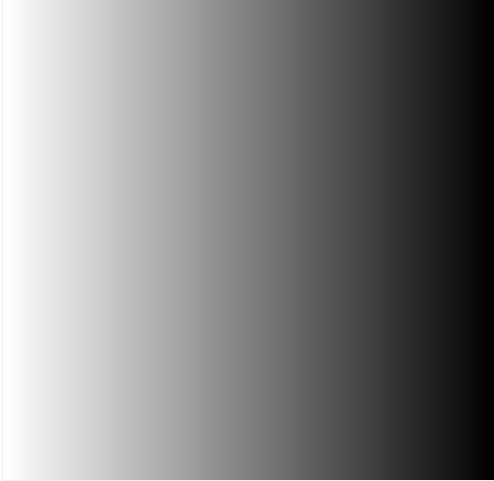}
\caption{Depth data}
\label{fig:depth}
\end{subfigure}
\begin{subfigure}{0.24\textwidth}
\includegraphics[width=\linewidth]{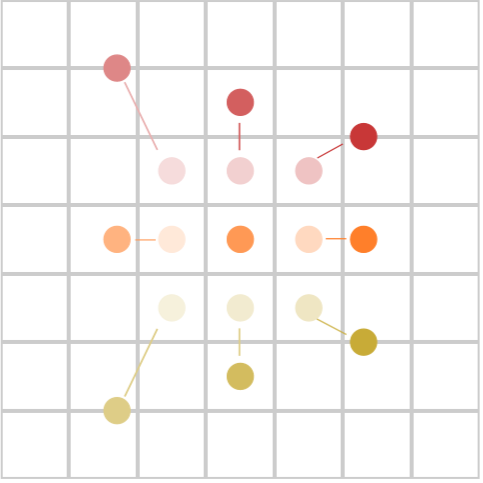}
\caption{Z-ACN}
\label{fig:guided}
\end{subfigure}
\caption{Effect of offset on a 3 $\times$ 3 kernel. a) shows a standard 2D convolution with dilation equals to 1. b) shows the offset computed from deformable convolution \cite{dai2017deformable}. c) is the available depth data. The represented figure shows a linear change in depth value. From left to right, the scene becomes deeper.  d) illustrates offset computed by Z-ACN which is adapted to depth.}
\label{fig:offset}
\end{figure}

\subsection{Back-projection on 3D Space}

Without loss of generality, we suppose that the camera fits the pinhole model. With RGB-D image, it is possible to back-project a 2D point $p(u_0, v_0)$ to 3D point, denoted as $P_0(X_0, Y_0, Z_0)$. In $p(u_0, v_0)$, instead of a fixed RF $\textbf{R(p)}$, we would like to propose a deformable RF by taking into account the geometric information.

Let us note $P_i = (X_i, Y_i, Z_i)$ the 3D points back-projected from $\textbf{R(p)}$. $i$ takes value from $0$ to $N\times N$ where $N \times N$ is the size of kernel. We extract the plane $\pi$ which passes through $P_0$ and fits the best to all $P_i$ :
\begin{equation}
\label{eq:plane}
\Vec{n} = \arg\min_{(n_1,n_2,n_3)} \sum_i ||n_1(X_i-X_0) + n_2(Y_i-Y_0) + n_3(Z_i-Z_0)||^2
\end{equation}
where $\Vec{n}=(n_1, n_2, n_3)$ is an approximation of the normal of the plane $\pi$ computed by singular value decomposition.

\subsection{3D Planar Grid}

For a 2D point $p(u_0, v_0)$, we may consider a conventional 2D convolution on image plane as realizing a planar convolution on a fronto-parallel plane on its back-projection $P_0(X_0, Y_0, Z_0)$ in 3D scene. By introducing the importance of depth, Z-ACN replaces the fronto-parallel plane by the new-defined plane $\pi$ that is adapted to the depth. In other words, Z-ACN computes a new planar and regular grid, denoted as $R_{3D}(P_0)$. $R_{3D}(P_0)$ is centered on $P_0$ and its regular shape is defined by an orthonormal basis $(\Vec{x'}, \Vec{y'})$ on $\pi$. We fix $\Vec{x'}$ horizontal ($\Vec{x'} = (\alpha, 0, \beta)$).  As $\Vec{x'}$ is on the plane $\pi$ defined by its normal $\Vec{n} = (n_1, n_2, n_3)$, we have : 
\begin{equation}
\begin{split}
    &\Vec{x'} \cdot \Vec{n} = 0, \quad ||\Vec{x'}||^2 = 1, \quad ||\Vec{n}||^2 = 1
\end{split}
\end{equation}

Analytically, we can compute the value for $\Vec{x}$ and $\Vec{y}=\Vec{n} \times \Vec{x'}$, such that:
\begin{equation}
\renewcommand\arraystretch{1.8}
\Vec{x'} = 
\begin{bmatrix}
\frac{n_3}{\sqrt{1 - n_2^2}} \\
0 \\
-\frac{n_1}{\sqrt{1 - n_2^2}}
\end{bmatrix},
\quad
\Vec{y'} = 
\begin{bmatrix}

-\frac{n_1n_2}{\sqrt{1 - n_2^2}} \\
\sqrt{1 - n_2^2}\\
-\frac{n_2n_3}{\sqrt{1 - n_2^2}}
\end{bmatrix}
\label{eq:basis}
\end{equation}

To conclude, $R_{3D}(P_0)$ is defined as : 

\begin{equation}
\label{eq:3dgrid}
R_{3D}(P_0) = a \Vec{x'} + b \Vec{y'}
\end{equation}
with $(a,b) \in (- k_u, 0 , k_u) \times (-k_v, 0 , k_v) $ where $(k_u, k_v)$ are scale factors. Their values will be discussed in section \ref{sec:scale}. 

The 3D grid on a depth-adapted plane guarantees Z-ACN to be a generic model. In the case when the plane is front-parallel, Z-ACN performs in the same way as a conventional CNN. Indeed, we have $\Vec{n}=(n_1, n_2, n_3) = (0,0,1)$. From Eq. \ref{eq:basis}, we have $\Vec{x'} = (1, 0, 0)$ and $\Vec{y'} = (0, 1, 0)$, which represent the regular shape for 3D grid. Thus, the projection on the image plane performs the same as a conventional or dilated convolution. Otherwise, being generic enables Z-ACN to be invariant to scale and rotation. 

\subsection{Z-ACN}

We denote $\textbf{R'(p)}$ the projection of $R_{3D}(P_0)$ on the image plane, which forms the Z-ACN : 
\begin{equation}
\begin{split}
   \textbf{y}(p) &= \sum_{p_n\in\textbf{R'(p)}} \textbf{w}(p)  \cdot \textbf{x}(p + p_n). \\
   & =  \sum_{p_n\in\textbf{R(p)}} \textbf{w}(p)  \cdot \textbf{x}(p + p_n + \Delta p_n)
\end{split}
\label{eq:zacn}
\end{equation}

Different from the conventional grid $\textbf{R(p)}$, the newly computed $\textbf{R'(p)}$ breaks the regular size and shape with the additional offset which contains more geometry information. 

\subsection{Scale Factor}
\label{sec:scale}
The scale factors $(k_u, k_v)$ are designed to be constant to guarantee the equal surface of 3D RF. In such a way, with the variance of depth, due to the perspective effect, the projected 2D RF on image plane will have different sizes.

The value of scale factors can be chosen in function of user's needs. In our application, we want Z-ACN performs the same as a conventional 2D convolution on a particular point $p(u_0,v_0)$ whose associated plane in Eq. \ref{eq:plane} is fronto-parallel $\{Z| Z = Z_{p}\}$. In other words : 
\begin{equation}
\sum_{p_n\in\textbf{R'(p)}} \textbf{w}(p_n)  \cdot \textbf{x}(p + p_n)  
=  \sum_{p_n\in\textbf{R(p)}} \textbf{w}(p_n)  \cdot \textbf{x}(p + p_n)
\end{equation}

By taking into account the dilation $\Delta d$ and the camera intrinsic parameters $(f_u, f_v)$ and by combing Eq. \ref{eq:2dgrid} and Eq. \ref{eq:3dgrid}, we have: 
\begin{equation}
\begin{split}
k_u = \Delta d \times \frac{Z_{p}}{f_u }  \\
k_v = \Delta d \times \frac{Z_{p}}{f_v }  \\
\label{scale}
\end{split}
\end{equation}

For any point with a deeper depth value than $Z_{p}$, the associated RF will be smaller. Otherwise, the associated RF will be equal or larger, which approves the fact of being adapted to scale.

\subsection{Understand Z-ACN}

Recent researches prove the performance of 2D CNN to understand the 3D scene. However, due to the regular grid, 2D CNN is more suitable for rigid objects where the deformation is minimal. To break this limit, there are 2 possibles ways. The first one is to augment the size of the dataset to contain all possible variations \cite{imagenet,silberman2012NYUV2} while the second is to augment the complexity, thus the ability, of network \cite{chen2017deeplab,He2016Residual,simonyan2014vgg}.

The latest advances in 3D sensors provide rich information about the geometry of 3D objects. 3D data can have different representations such as voxel, point cloud, and multi-view image. Studies \cite{maturana2015voxnet,Chen_2017MultiView,pointcloud} show the impact of different representation on the performance. However, these approaches based on 3D data suffer from high computation complexity.

\begin{figure}[t]
\centering
\includegraphics[width=\linewidth,keepaspectratio]{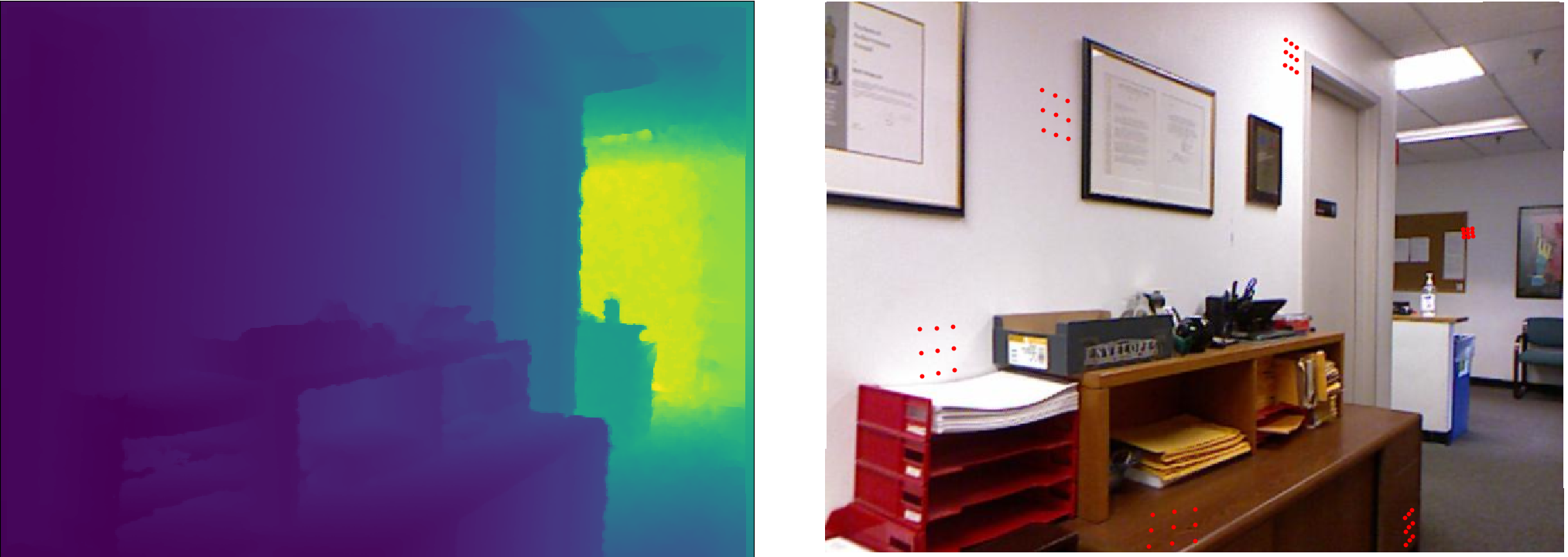}
\caption{The left figure shows the depth map. Instead of conventional 2D regular RF, Z-ACN is a generic model that takes into account the geometry. Red points on the right hand represent our RF for a $3\times 3$ kernel. As shown in the image, our model enables a modification on the convolution grid that describes better the geometry in a 2D image, which helps to be invariant to rotation. With the variation of depth, the surface convolution grid changes as well, which helps to be invariant to scale.}
\label{fig:Z-ACN}
\end{figure}

RGB-D images seem to be the most accessible and light data that articulates both 2D and 3D advantages. Z-ACN takes this particularity to include the depth into the convolution by adjusting the 2D convolutional grid. This pattern is integrated into Eq. \ref{eq}. To get a better understanding of Z-ACN, Fig. \ref{fig:Z-ACN} shows the depth-adapted 2D grid of a given input neuron (the center). In the case of conventional CNN, the shape of the grid is fixed as regular, which has difficulties to deal with 3D information. With the Z-ACN, the grid is adjusted to geometry. As shown in Fig. \ref{fig:Z-ACN}, the receptive field for a nearer point is larger than that of a farther point. Receptive fields on the same plane also have different shapes that are adapted to the camera-projection effect. These patterns increase 2D CNN's performance to adapt to potential transformation without complicating the network nor augmenting the size of the dataset.

\section{Experiment Work}

As a generic model, Z-ACN can apply to all applications such as classification, segmentation, object detection, etc. In this paper, we evaluate our model on the problem of semantic segmentation. For any given CNN, we follow the same configuration on loss function and optimizer and replace the only convolution by Z-ACN operation. The whole work is realized under the Pytorch framework. We use the official deformable convolution from the torchvision package. The scale factor in Eq. \ref{scale} is computed with the mean of the input depth map. We repeat all experiments three times on an Nvidia 2080 Ti GPU and report the average model performance.

Experiments are realized with the NYUv2 dataset \cite{silberman2012NYUV2}. We take 1,449 RGB-D images with pixel-wise labels. We split them into 795 training images and 654 testing images. For labels, we follow respectively the 13-class settings and the 37-class settings. Note that Z-ACN works with RGB-D images, but the input of CNN models remains RGB. The depth information is only introduced in the guided deformable convolution. In other words, adapted CNN models only extract features from the RGB image, which is the same as the initial CNN model. Only the sample position is guided by the depth. To make difference from classical RGB input and classical RGB-D or RGB + HHA input, we denote RGB(D) as the input of Z-ACN.

\subsection{Integrating Depth in RGB Convolution}
\label{sec:integrate}

As the Z-ACN is invariant to scale and rotation, in this section, we want to show that Z-ACN should achieve better results than conventional convolution in existing architecture. We choose U-Net \cite{ronneberger2015unet} as our baseline. We train U-Net with respectively RGB input, RGB-D input, and RGB(D) input. All models are trained from scratch with the NYUv2 dataset following 13-class settings. We use conventional cross-entropy as loss function, SGD optimizer with initial learning rate 0.0001, momentum 0.99, and batch size 1. 

We evaluate the performance by regarding common metrics: overall accuracy, mean accuracy, mean intersection over union, and frequent weighted intersection over union. 
Overall Accuracy (Acc) stands for the proportion of correctly predicted pixels in the whole image. 
Mean Accuracy (mAcc) further analyzes the accuracy averaged over all the classes. 
Intersection over Union (IoU) studies the proportion of overlap area between the predicted segmentation and the ground truth divided by the area union and averaged over all the classes. 
Frequency Weighted Intersection over Union (fwIoU) further analyzes the IoU weighted by the total pixel ratio of each class. 

Mathematically, suppose that we have $s_i$ the number of pixels with the ground truth class $i$. We can compute the total number of all pixels: $s = \sum_i s_i$. $n_{ij}$ denotes the number of pixels with ground truth class $i$ and predicted as class $j$, $n_c$ denotes the number of total classes. The model is evaluated by: 
\begin{itemize}
    \item Overall Acc: $Acc = \sum_i \frac{n_{ii}}{s}$ 
    \item Mean ACC: $mAcc = \frac{1}{n_c}  \sum_i \frac{n_{ii}}{s}$ 
    \item mean Intersection over Union: $mIoU = \frac{1}{n_c}  \sum_i \frac{n_{ii}}{s_i + \sum_j n_{ji} - n_{ii}}$ 
    \item Frequency Weighted Intersection over Union: $fwIoU = \frac{1}{s}  \sum_i s_i\frac{n_{ii}}{s_i + \sum_j n_{ji} - n_{ii}}$ 
\end{itemize}

The results of our experiments are summarised in Table \ref{unet}. We can observe that the network with additional depth information achieves a better result than the network with RGB only images. But the result with RGB(D) input outperforms that with RGB-D input. Compared to RGB and RGB-D input, the improvement with Z-ACN was +3.2\% and +2.8\% for the pixel-wise accuracy; +3.9\% and +5.1\% for class accuracy; +4.6\% and +3.2\% for mean Intersection of Union; and +2.8\% and +2.1\% for Frequent Weighted Intersection of Union. Note that the number of parameters is the same to extract features from the RGB image, which is also the case of Z-ACN. But for the network with RGB-D input, it requires a slightly higher number of parameters due to the input size. Nevertheless, it is still outperformed by our model. This result proves that Z-ACN integrates depth information more effectively in the RGB CNN.

\begin{table}[ht]
\centering
\begin{tabular}[ht]{ m{1.5cm}| m{1.5cm}|m{1.5cm} m{1.5cm} m{1.5cm} }
\hline
\multicolumn{2}{c|}{NYUv2 13 class} & RGB  & RGB-D &  RGB(D)  \\
\hline
\multirow{4}{*}[-3pt]{U-Net\cite{ronneberger2015unet}} 
& Acc (\%)&  52.4& 52.8  &  \textbf{55.6}\\
& mAcc (\%)&  40.4 &  39.2  & \textbf{44.3}\\
& mIoU (\%)&  27.2 & 28.6  & \textbf{31.8} \\
& fwIoU (\%)&  36.9 &  37.6 & \textbf{39.7}\\
\hline
\end{tabular}
\caption{Comparison with different inputs on U-Net on NYUv2 test set following 13-class setting. Networks are trained from scratch. We test respectively U-Net with RGB input, with RGB-D input and with RGB(D) inputs. All evaluations perform at 640 x 480 resolution. We show that Z-ACN attends better result than other inputs.}
\label{unet}
\end{table}

\subsection{Comparison with the State-of-the-art}

As the Z-ACN is invariant to scale and rotation around the X and the Y axis from the camera coordinate system, we want to show that Z-ACN should achieve similar results with less learning parameters. In other words, our model should enable a CNN model with fewer feature channels to have similar performance. 
To the best of our knowledge, \cite{chu2018surfconv} is the latest research working in the same direction with RGB-D images. 

To prove this idea, we adopt a modified ResNet-18 \cite{He2016Residual} as our backbone. We replace the conventional convolution by Z-ACN. We change the size of all convolutional kernels to be $3\times 3$. We reduce the number of feature channels by $2^4$. The input of our network is classical the RGB(D) image. We use the skip-connected fully convolutional architecture \cite{Long2015FCN}. This network is trained from scratch. We train the new model with the NYUv2 dataset following 37-class settings for a complex scene analyzing. We randomly sample 20\% rooms from the training set as the validation set. The model is trained with the logarithm loss function.

We compare Z-ACN with 3D representation such as PointNet\cite{Qi2017pointnet}, Conv3D \cite{tchapmi2017segcloud,song2017semantic} and 2D representation such as DeformCNN\cite{dai2017deformable} and SurfConv \cite{chu2018surfconv}. Conv3D and PointNet use the hole-filled dense depth map provided by the dataset to create voxel input. For PointNet, the source code is used to uses RGB plus gravity-aligned point cloud. The recommended configuration \cite{Qi2017pointnet} is used to randomly sample points. The sample number is set to be 25k. For Conv3D, the SSCNet architecture \cite{tchapmi2017segcloud} is used and is trained with flipped - TSDF and RGB. The resolution is reduced to be $240\times144\times240$ voxel grid. For DeformCNN, RGB images and HHA images are chosen as input for a fair comparison.  For SurfConv, we compare with both the 1-level model and the recommended 4-level model \cite{chu2018surfconv} (the 4-level model requires a resampling on the input image to be adapted to the different levels of depth). For all the above-mentioned models, we follow the same configuration and learning settings as discussed in \cite{chu2018surfconv}. Thus we compare directly with the presented results.

\begin{table}[ht]
\centering
\begin{tabular}[ht]{l|c|c |c |c c}
\hline
NYUv2 37 class & Input & \# of param & mIoU (\%) & Acc (\%)\\
\hline
PointNet\cite{Qi2017pointnet} & voxel + RGB & 1675k & 6.9 & 47.4 \\
\hline

Conv3D \cite{tchapmi2017segcloud,song2017semantic} & voxel + RGB & 241k & 13.2 & 49.9 \\
\hline
DeformCNN\cite{dai2017deformable} & HHA + RGB & 101k & 12.8 & 55.1 \\
\hline
SurfConv1\cite{chu2018surfconv} & HHA + RGB & 65k & 12.3 & 53.7 \\
\hline
SurfConv4\cite{chu2018surfconv} & HHA + RGB & 65k & 13.1 & 53.5 \\
\hline
Z-ACN (ours) & RGB(D) & 65k &\textbf{13.5} & \textbf{57.2} \\
\hline
\end{tabular}
\caption{Comparison with different models with NYUv2 test set following 37-class settings.  All results except ours are extracted from \cite{chu2018surfconv}. Our model is trained from scratch with the same settings.}
\label{compscale}
\end{table}

\begin{figure}[ht]
\centering
\includegraphics[width=\linewidth,keepaspectratio]{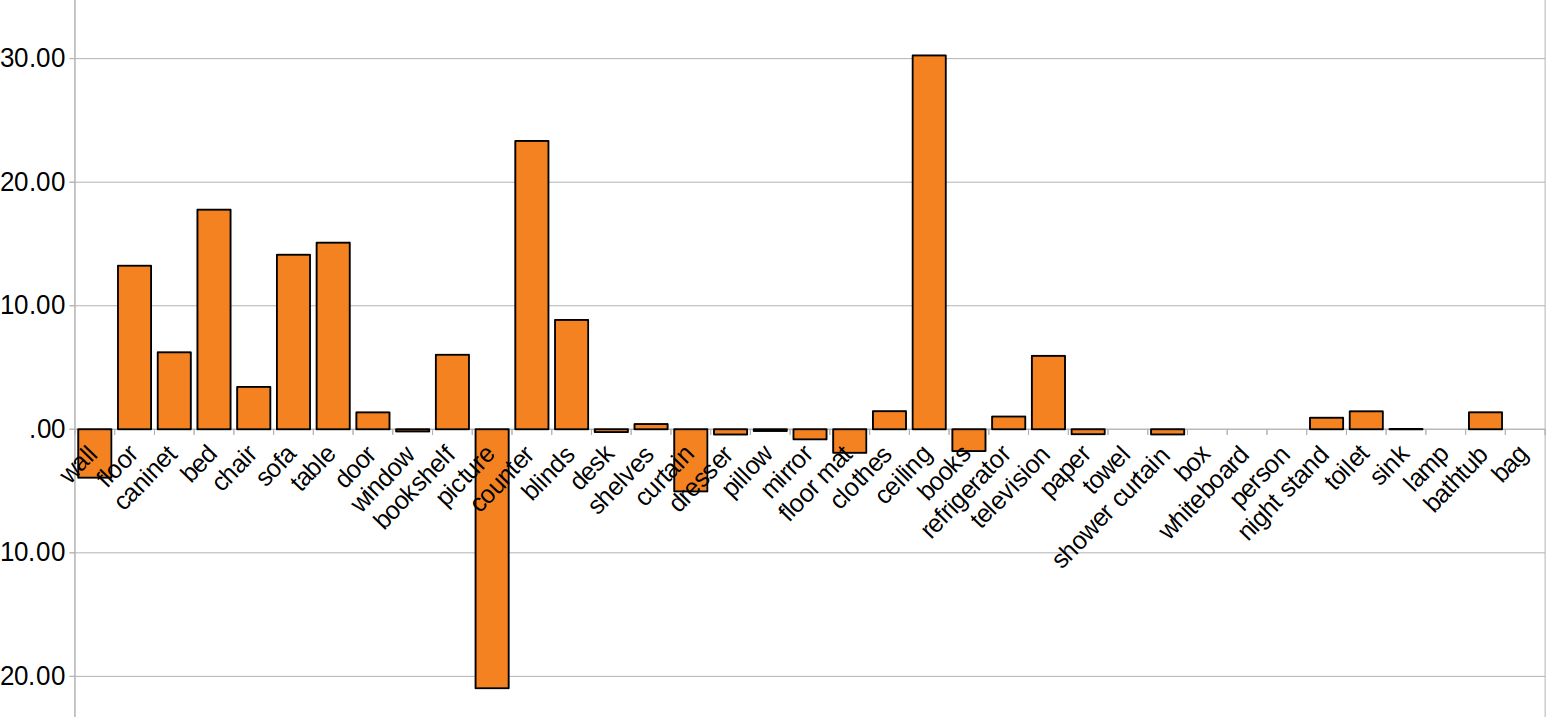}
\caption{Illustration of average improved percentage of per-class IoU. We compare Z-ACN with SurfConv single-level, with the exact same CNN model. Models are trained from scratch. On NYUv2, we improve 25/37 classes with 6.1\% mean IoU increment.}
\label{fig:comp}
\end{figure}

As shown in Table \ref{compscale}, Z-ACN achieves better results compared to all other methods:
\begin{itemize}
    \item Compared to PointNet, Z-ACN uses only 4\% of its number of parameters but achieves +7\% on mIoU and + 10\% on Acc.
    \item Compared to Conv3D, Z-ACN uses less than 30\% of its number of parameters to achieves close but better results on mIoU and +7\% on Acc.
    \item Compared to DeformCNN which also adds offset to the convolution, Z-ACN uses only 65\% of its parameters as the offset in Z-ACN does not require gradient during back-propagation. Z-ACN still achieves close but better results on mIoU and +2\% on Acc.
    \item Compared to SurfConv, with 1-level configuration (the same size of input data) and the same model (FCN + ResNet-18), Z-ACN achieves +1\% on mIoU and + 4\% on Acc. Compared to 4-level which resamples input data, Z-ACN remains to be better in both mIoU and Acc.  
\end{itemize}

Fig. \ref{fig:comp} illustrates the average improvement of per-class IoU between Z-ACN and SurfConv1. As our model adapts convolution to a local plane, it should be sensitive to depth differences. We can observe that Z-ACN achieves significantly better results on objects with large sizes such as floor, bed, sofa, table, counter, and ceiling. Recognizably, large-size objects can be easily distinguished from other objects because commonly they don't share the same 3D plane. This result meets our expectations. Wall is also a large-size object, but as shown in Fig. \ref{fig:Z-ACN}, there might be other objects such as pictures on the same plane which adds ambiguities to our model.

Different from 3D representation, the Z-ACN remains a light-weight 2D CNN which consumes significantly less memory. Z-ACN performs also better than other state-of-the-art models with RGB-only input. The result validates the effectiveness of integrating depth information in convolution by adapting the sampling position.

\subsection{Adapting to Existing Model}
Previous results show that the memory requirement of our network is low but it can still achieve better results compared to memory-consuming approaches. As Z-ACN does not add extra learning parameters to the existing model, we can take advantage of heavier models to further improve its performance. 

\begin{table}[ht]
\centering
\begin{tabular}[ht]{ m{1.5cm} | m{1.5cm} m{1.5cm} m{1.5cm} m{1.5cm} m{1.5cm} }
\hline
\multicolumn{1}{c|}{NYUv2 13 class} & Acc (\%)  & mAcc (\%) &  mIoU (\%) & fwIoU (\%)  \\
\hline
Baseline & 63.9 & 49.0  & 36.7 & 48.6 \\
\cline{1-1} Z-ACN & \textbf{69.4} & \textbf{56.5}  & \textbf{43.8} & \textbf{55.1} \\
\hline
\hline
\multicolumn{1}{c|}{NYUv2 37 class} & Acc (\%)  & mAcc (\%) &  mIoU (\%) & fwIoU (\%)  \\
\hline
Baseline & 70.0 & 34.3  & 25.7 & 54.9 \\
\cline{1-1} Z-ACN & \textbf{73.5} & \textbf{36.8}  & \textbf{28.4} & \textbf{58.8} \\
\hline
\end{tabular}
\caption{Comparison between Deeplab with VGG-16 as encoder and Z-ACN on NYUv2 test set following 13-class setting and 37-class setting. Networks are trained from scratch. We show that our model attends better result than initial model.}
\label{deeplab}
\end{table}

We choose VGG-16 \cite{simonyan2014vgg} as our CNN network which is widely used in the NYUv2 dataset. We use Deeplab \cite{chen2017deeplab} with VGG-16 as the baseline. This model requires 20520k parameters. A quantitative comparison between the initial model and Z-ACN is summarized in Table \ref{deeplab}. We use common metrics and configurations presented in Section \ref{sec:integrate}. Both models are trained from scratch with the NYUv2 dataset following respectively  13-class settings and 37-class settings.

Table \ref{deeplab} illustrates that with more parameters, CNN can achieve better results. It also shows that our model Z-ACN can improve the performance of a memory-consuming model as well, without adding extra parameters.

\section{Conclusions}

Z-ACN is a novel and generic model to include geometry data provided by RGB-D images in 2D CNN. When the depth is unknown, e.g. RGB images, or when the scene is planar and the image plane is parallel to this scene, Z-ACN performs in the same way as dilated or conventional convolution which helps to be invariant to scale. In other cases, it provides a novel method to adapt the receptive field to the geometry, which is invariant to scale and rotation around the X and the Y axis from the camera coordinate system. This design helps to improve the performance of existing CNN to attend the same result by reducing the computation cost, or to achieve better results without complicating the network architecture nor adding extra training parameters.

We demonstrated the effectiveness of Z-ACN on semantic segmentation task. Results are trained from scratch. In future works, we will figure out how to adapt Z-ACN to existing pre-trained models and how to adapt the geometry in the pooling layer. We will also try to extend the application to other popular tasks like normal or depth estimation, instance segmentation, or even pass from RGB-D image to 3D dataset. 

\section*{Acknowledgements}

We gratefully acknowledge Zhuyun Zhou for her support and proofread. We also thank Clara Fernandez-Labrador, Marc Blanchon, and Zhihao Chen for the discussion. This research is supported by the French National Research Agency through ANR CLARA (ANR-18-CE33-0004) and financed by the French Conseil R\'egional de Bourgogne-Franche-Comt\'e.  

\bibliographystyle{splncs}
\bibliography{egbib}

\begin{thebibliography}{10}

\bibitem{chen2017deeplab}
Chen, L.C., Papandreou, G., Kokkinos, I., Murphy, K., Yuille, A.L.:
\newblock Deeplab: Semantic image segmentation with deep convolutional nets,
  atrous convolution, and fully connected crfs.
\newblock IEEE transactions on pattern analysis and machine intelligence
  (PAMI). (2017)

\bibitem{Long2015FCN}
Long, J., Shelhamer, E., Darrell, T.:
\newblock Fully convolutional networks for semantic segmentation.
\newblock In: The IEEE Conference on Computer Vision and Pattern Recognition
  (CVPR). (2015)

\bibitem{chen2017Atrous}
Chen, L.C., Papandreou, G., Schroff, F., Adam, H.:
\newblock Rethinking atrous convolution for semantic image segmentation.
\newblock arXiv:1706.05587 (2017)

\bibitem{Eigen2015MultiScale}
Eigen, D., Fergus, R.:
\newblock Predicting depth, surface normals and semantic labels with a common
  multi-scale convolutional architecture.
\newblock In: International Conference on Computer Vision (ICCV). (2015)

\bibitem{girshick2015fastrcnn}
Girshick, R.:
\newblock Fast r-cnn.
\newblock In: Proceedings of the IEEE international conference on computer
  vision (ICCV). (2015)

\bibitem{girshick2014rcnn}
Girshick, R., Donahue, J., Darrell, T., Malik, J.:
\newblock Rich feature hierarchies for accurate object detection and semantic
  segmentation.
\newblock In: Proceedings of the IEEE conference on computer vision and pattern
  recognition (CVPR). (2014)

\bibitem{he2017maskrcnn}
He, K., Gkioxari, G., Doll{\'a}r, P., Girshick, R.:
\newblock Mask r-cnn.
\newblock In: Proceedings of the IEEE international conference on computer
  vision (ICCV). (2017)

\bibitem{hazirbas2016fusenet}
Hazirbas, C., Ma, L., Domokos, C., Cremers, D.:
\newblock Fusenet: Incorporating depth into semantic segmentation via
  fusion-based cnn architecture.
\newblock In: Asian conference on computer vision (ACCV). (2016)

\bibitem{hu2019acnet}
Hu, X., Yang, K., Fei, L., Wang, K.:
\newblock Acnet: Attention based network to exploit complementary features for
  rgbd semantic segmentation.
\newblock In: 2019 IEEE International Conference on Image Processing (ICIP),
  IEEE (2019)

\bibitem{Lin2017RGBDCascaded}
Lin, D., Chen, G., Cohen-Or, D., Heng, P.A., Huang, H.:
\newblock Cascaded feature network for semantic segmentation of rgb-d images.
\newblock In: The IEEE International Conference on Computer Vision (ICCV).
  (2017)

\bibitem{park2017rdfnet}
Park, S.J., Hong, K.S., Lee, S.:
\newblock Rdfnet: Rgb-d multi-level residual feature fusion for indoor semantic
  segmentation.
\newblock In: Proceedings of the IEEE International Conference on Computer
  Vision (ICCV). (2017)

\bibitem{wang2016RGBD}
Wang, J., Wang, Z., Tao, D., See, S., Wang, G.:
\newblock Learning common and specific features for rgb-d semantic segmentation
  with deconvolutional networks.
\newblock In: European Conference on Computer Vision (ECCV), Springer (2016)

\bibitem{chen2018parallel}
Chen, L.C., Zhu, Y., Papandreou, G., Schroff, F., Adam, H.:
\newblock Encoder-decoder with atrous separable convolution for semantic image
  segmentation.
\newblock In: Proceedings of the European conference on computer vision (ECCV).
  (2018)

\bibitem{yu2015Cascade}
Yu, F., Koltun, V.:
\newblock Multi-scale context aggregation by dilated convolutions.
\newblock arXiv preprint arXiv:1511.07122 (2015)

\bibitem{chen2016scale1}
Chen, L.C., Yang, Y., Wang, J., Xu, W., Yuille, A.L.:
\newblock Attention to scale: Scale-aware semantic image segmentation.
\newblock In: Proceedings of the IEEE conference on computer vision and pattern
  recognition (CVPR). (2016)

\bibitem{xia2016scale2}
Xia, F., Wang, P., Chen, L.C., Yuille, A.L.:
\newblock Zoom better to see clearer: Human and object parsing with
  hierarchical auto-zoom net.
\newblock In: European Conference on Computer Vision (ECCV). (2016)

\bibitem{li2019pyramid}
Li, Y., Chen, Y., Wang, N., Zhang, Z.:
\newblock Scale-aware trident networks for object detection.
\newblock In: Proceedings of the IEEE International Conference on Computer
  Vision (ICCV). (2019)

\bibitem{lin2017Pyramid}
Lin, T.Y., Doll{\'a}r, P., Girshick, R., He, K., Hariharan, B., Belongie, S.:
\newblock Feature pyramid networks for object detection.
\newblock In: Proceedings of the IEEE conference on computer vision and pattern
  recognition (CVPR). (2017)

\bibitem{Zhao2017Pyramid}
Zhao, H., Shi, J., Qi, X., Wang, X., Jia, J.:
\newblock Pyramid scene parsing network.
\newblock In: The IEEE Conference on Computer Vision and Pattern Recognition
  (CVPR). (2017)

\bibitem{jaderberg2015spatial}
Jaderberg, M., Simonyan, K., Zisserman, A.,  et~al.:
\newblock Spatial transformer networks.
\newblock In: Advances in neural information processing systems (NIPS). (2015)

\bibitem{dai2017deformable}
Dai, J., Qi, H., Xiong, Y., Li, Y., Zhang, G., Hu, H., Wei, Y.:
\newblock Deformable convolutional networks.
\newblock In: Proceedings of the IEEE international conference on computer
  vision (ICCV). (2017)

\bibitem{Zhang2019locationdeform}
Zhang, C., Kim, J.:
\newblock Object detection with location-aware deformable convolution and
  backward attention filtering.
\newblock In: The IEEE Conference on Computer Vision and Pattern Recognition
  (CVPR). (2019)

\bibitem{tian2020tdan}
Tian, Y., Zhang, Y., Fu, Y., Xu, C.:
\newblock Tdan: Temporally-deformable alignment network for video
  super-resolution.
\newblock In: Proceedings of the IEEE/CVF Conference on Computer Vision and
  Pattern Recognition (CVPR). (2020)

\bibitem{wang2019edvr}
Wang, X., Chan, K.C., Yu, K., Dong, C., Change~Loy, C.:
\newblock Edvr: Video restoration with enhanced deformable convolutional
  networks.
\newblock In: Proceedings of the IEEE Conference on Computer Vision and Pattern
  Recognition Workshops (CVPRW). (2019)

\bibitem{eder2019mapped}
Eder, M., Price, T., Vu, T., Bapat, A., Frahm, J.M.:
\newblock Mapped convolutions.
\newblock arXiv preprint arXiv:1906.11096 (2019)

\bibitem{tateno2018distortion}
Tateno, K., Navab, N., Tombari, F.:
\newblock Distortion-aware convolutional filters for dense prediction in
  panoramic images.
\newblock In: Proceedings of the European Conference on Computer Vision (ECCV).
  (2018)

\bibitem{coors2018spherenet}
Coors, B., Paul~Condurache, A., Geiger, A.:
\newblock Spherenet: Learning spherical representations for detection and
  classification in omnidirectional images.
\newblock In: Proceedings of the European Conference on Computer Vision (ECCV).
  (2018)  518--533

\bibitem{3DConvLanding}
{Maturana}, D., {Scherer}, S.:
\newblock 3d convolutional neural networks for landing zone detection from
  lidar.
\newblock In: 2015 IEEE International Conference on Robotics and Automation
  (ICRA). (2015)

\bibitem{maturana2015voxnet}
Maturana, D., Scherer, S.:
\newblock Voxnet: A 3d convolutional neural network for real-time object
  recognition.
\newblock In: 2015 IEEE/RSJ International Conference on Intelligent Robots and
  Systems (IROS), IEEE (2015)

\bibitem{wu20153d}
Wu, Z., Song, S., Khosla, A., Yu, F., Zhang, L., Tang, X., Xiao, J.:
\newblock 3d shapenets: A deep representation for volumetric shapes.
\newblock In: Proceedings of the IEEE conference on computer vision and pattern
  recognition (CVPR). (2015)

\bibitem{Qi2017pointnet}
Qi, C.R., Su, H., Mo, K., Guibas, L.J.:
\newblock Pointnet: Deep learning on point sets for 3d classification and
  segmentation.
\newblock In: The IEEE Conference on Computer Vision and Pattern Recognition
  (CVPR). (2017)

\bibitem{NIPS2017Pointnet}
Qi, C.R., Yi, L., Su, H., Guibas, L.J.:
\newblock Pointnet++: Deep hierarchical feature learning on point sets in a
  metric space.
\newblock In Guyon, I., Luxburg, U.V., Bengio, S., Wallach, H., Fergus, R.,
  Vishwanathan, S., Garnett, R., eds.: Advances in Neural Information
  Processing Systems 30.
\newblock Curran Associates, Inc. (2017)

\bibitem{li2018pointcnn}
Li, Y., Bu, R., Sun, M., Wu, W., Di, X., Chen, B.:
\newblock Pointcnn: Convolution on x-transformed points.
\newblock In: Advances in neural information processing systems. (2018)

\bibitem{Thomas2019kpconv}
Thomas, H., Qi, C.R., Deschaud, J.E., Marcotegui, B., Goulette, F., Guibas,
  L.J.:
\newblock Kpconv: Flexible and deformable convolution for point clouds.
\newblock In: The IEEE International Conference on Computer Vision (ICCV).
  (2019)

\bibitem{liu2019relation}
Liu, Y., Fan, B., Xiang, S., Pan, C.:
\newblock Relation-shape convolutional neural network for point cloud analysis.
\newblock In: Proceedings of the IEEE Conference on Computer Vision and Pattern
  Recognition. (2019)

\bibitem{liu2019densepoint}
Liu, Y., Fan, B., Meng, G., Lu, J., Xiang, S., Pan, C.:
\newblock Densepoint: Learning densely contextual representation for efficient
  point cloud processing.
\newblock In: Proceedings of the IEEE International Conference on Computer
  Vision. (2019)

\bibitem{tchapmi2017segcloud}
Tchapmi, L., Choy, C., Armeni, I., Gwak, J., Savarese, S.:
\newblock Segcloud: Semantic segmentation of 3d point clouds.
\newblock In: 2017 international conference on 3D vision (3DV), IEEE (2017)

\bibitem{Chen_2017MultiView}
Chen, X., Ma, H., Wan, J., Li, B., Xia, T.:
\newblock Multi-view 3d object detection network for autonomous driving.
\newblock In: The IEEE Conference on Computer Vision and Pattern Recognition
  (CVPR). (2017)

\bibitem{Ge20173Dpose}
Ge, L., Liang, H., Yuan, J., Thalmann, D.:
\newblock 3d convolutional neural networks for efficient and robust hand pose
  estimation from single depth images.
\newblock In: The IEEE Conference on Computer Vision and Pattern Recognition
  (CVPR). (2017)

\bibitem{3DFCN}
Li, B., Zhang, T., Xia, T.:
\newblock Vehicle detection from 3d lidar using fully convolutional network.
\newblock arXiv preprint arXiv:1608.07916 (2016)

\bibitem{Qi2016VolMultiView}
Qi, C.R., Su, H., Niessner, M., Dai, A., Yan, M., Guibas, L.J.:
\newblock Volumetric and multi-view cnns for object classification on 3d data.
\newblock In: The IEEE Conference on Computer Vision and Pattern Recognition
  (CVPR). (2016)

\bibitem{Song2016shape}
Song, S., Xiao, J.:
\newblock Deep sliding shapes for amodal 3d object detection in rgb-d images.
\newblock In: The IEEE Conference on Computer Vision and Pattern Recognition
  (CVPR). (2016)

\bibitem{song2017semantic}
Song, S., Yu, F., Zeng, A., Chang, A.X., Savva, M., Funkhouser, T.:
\newblock Semantic scene completion from a single depth image.
\newblock In: Proceedings of the IEEE Conference on Computer Vision and Pattern
  Recognition (CVPR). (2017)

\bibitem{qi2018frustum}
Qi, C.R., Liu, W., Wu, C., Su, H., Guibas, L.J.:
\newblock Frustum pointnets for 3d object detection from rgb-d data.
\newblock In: Proceedings of the IEEE conference on computer vision and pattern
  recognition (CVPR). (2018)

\bibitem{Tang2019frustumplus}
Tang, Y.S., Lee, G.H.:
\newblock Transferable semi-supervised 3d object detection from rgb-d data.
\newblock In: The IEEE International Conference on Computer Vision (ICCV).
  (2019)

\bibitem{Wang2018DCNN}
Wang, W., Neumann, U.:
\newblock Depth-aware cnn for rgb-d segmentation.
\newblock In: The European Conference on Computer Vision (ECCV). (2018)

\bibitem{chu2018surfconv}
Chu, H., Ma, W.C., Kundu, K., Urtasun, R., Fidler, S.:
\newblock Surfconv: Bridging 3d and 2d convolution for rgbd images.
\newblock In: Proceedings of the IEEE Conference on Computer Vision and Pattern
  Recognition (CVPR). (2018)

\bibitem{imagenet}
Deng, J., Dong, W., Socher, R., Li, L.J., Li, K., Fei-Fei, L.:
\newblock Imagenet: A large-scale hierarchical image database.
\newblock In: 2009 IEEE conference on computer vision and pattern recognition
  (CVPR). (2009)

\bibitem{silberman2012NYUV2}
Silberman, N., Hoiem, D., Kohli, P., Fergus, R.:
\newblock Indoor segmentation and support inference from rgbd images.
\newblock In: European conference on computer vision (ECCV). (2012)

\bibitem{He2016Residual}
He, K., Zhang, X., Ren, S., Sun, J.:
\newblock Deep residual learning for image recognition.
\newblock In: The IEEE Conference on Computer Vision and Pattern Recognition
  (CVPR). (2016)

\bibitem{simonyan2014vgg}
Simonyan, K., Zisserman, A.:
\newblock Very deep convolutional networks for large-scale image recognition.
\newblock arXiv preprint arXiv:1409.1556 (2014)

\bibitem{pointcloud}
{Jing Huang}, {Suya You}:
\newblock Point cloud labeling using 3d convolutional neural network.
\newblock In: 2016 23rd International Conference on Pattern Recognition (ICPR).
  (2016)

\bibitem{ronneberger2015unet}
Ronneberger, O., Fischer, P., Brox, T.:
\newblock U-net: Convolutional networks for biomedical image segmentation.
\newblock In: International Conference on Medical image computing and
  computer-assisted intervention (MICCAI). (2015)

\end{thebibliography}

\end{document}